\documentclass[a4paper,11pt]{article}
\usepackage[paper=a4paper,top=1.5cm,left=2cm,right=2cm, foot=1cm,bottom=1.5cm, total={17cm,25.7cm}]{geometry}

\usepackage[utf8]{inputenc}
\usepackage{amsmath,amssymb,amsthm,mathtools,mathrsfs,stmaryrd,titletoc}
\usepackage[scaled=0.92]{helvet}
\usepackage{graphicx}
\usepackage{subcaption}
\usepackage{authblk}
\usepackage[bookmarks=true,colorlinks=true,linkcolor=blue,citecolor=blue]{hyperref}

\usepackage{tabularx}
\usepackage{booktabs}
\usepackage{multirow}

\usepackage{algorithm}
\usepackage{algorithmicx}
\usepackage{algpseudocode}
\usepackage{indentfirst}

\usepackage{siunitx}
\usepackage[english]{babel}
\usepackage[usenames]{color}
\usepackage{graphicx}
\graphicspath{{./images/}}

\usepackage{natbib}

\setcitestyle{authoryear,open={(},close={)}}

\usepackage[title, titletoc, toc]{appendix}
\usepackage[capitalise]{cleveref}
\definecolor{mygreen}{rgb}{0,0.6,0}
\definecolor{darkgray}{rgb}{0.95,0.95,0.95}
\definecolor{lightgray}{gray}{0.9}
\definecolor{myblue}{rgb}{0.01, 0.31, 0.59}
\crefname{figure}{Fig.}{Figs.}

\title{A novel activity pattern generation incorporating deep learning for transport demand models}
\date{} 
\author[a]{Danh T. Phan\thanks{Corresponding author: the.phan@monash.edu, hai.vu@monash.edu}}
\author[a]{Hai L. Vu}
\affil[a]{\footnotesize \textit{Department of Civil Engineering, Monash University, Clayton 3800, VIC, Australia}}

\begin{document}
\maketitle
\begin{abstract}
Activity generation plays an important role in activity-based demand modelling systems. While machine learning, especially deep learning, has been increasingly used for mode choice and traffic flow prediction, much less research exploiting the advantage of deep learning for activity generation tasks. This paper proposes a novel activity pattern generation framework by incorporating deep learning with travel domain knowledge. We model each activity schedule as one primary activity tour and several secondary activity tours. We then develop different deep neural networks with entity embedding and random forest models to classify activity type, as well as to predict activity times. The proposed framework can capture the activity patterns for individuals in \textit{both} training and validation sets. Results show high accuracy for the start time and end time of work and school activities. The framework also replicates the start time patterns of stop-before and stop-after primary work activity well. This provides a promising direction to deploy advanced machine learning methods to generate more reliable activity-travel patterns for transport demand systems and their applications.
\end{abstract}


\section{Introduction}\label{intro}

\begin{figure}[!h]
  \centering
  \includegraphics[scale=0.75]{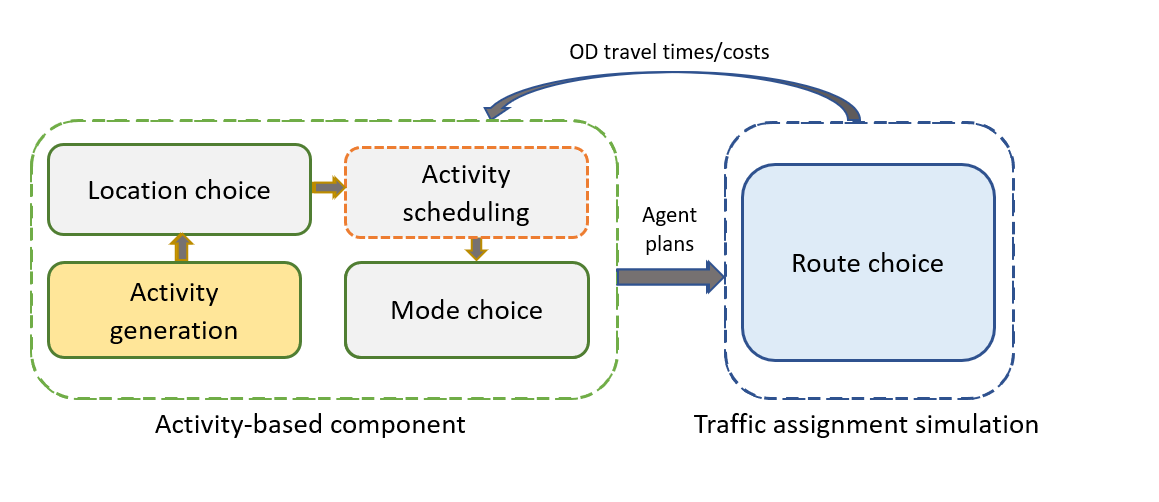}
  \caption{The simplified architecture of an agent-based system}
  \label{fig:agent-based-system}
\end{figure}

Activity-based travel demand models have been increasingly used to support transport planners to forecast mobility patterns \citep{lr-castiglione2015activity}. These demand models aim to describe the process of how individuals plan and schedule their daily activities that affect travel demand \citep{lr-axhausen1992activity}. Specifically, it models and imitates the individual activity decision of what, when, how long, where, and with whom involved. Activity-based models usually include activity generation, activity location choice, and mode choice modules. An activity generation module is responsible for answering what, when and how long the activity will be carried out; a location choice module addresses the question of where the activity occurred; a mode choice module is used to model how individuals travel between activities. Some activity-based models may additionally have an activity scheduling module that simulates the activity scheduling process and interaction among individuals or members in households. An activity-based model is often integrated with a traffic assignment simulation to form an agent-based system \citep{mobitopp-briem2019creating}, as shown in \cref{fig:agent-based-system}. The outcomes of activity-based models are individuals' activity schedules or plans. These schedules are the demand input for agent-based simulations. Being the first module, the activity generation plays a crucial role in developing an accurate and realistic transport demand model.

Two popular approaches for activity generation are utility-based and rule-based models \citep{lr-rasouli2014activity}. The utility-based models are based on an econometric assumption that individuals select a travel pattern alternative that maximises their utilities. Discrete choice methods are often used to estimate and select the choice with the highest utility among different alternatives. While discrete choice methods could effectively estimate output variables having a small number of discrete values like travel modes, these methods could face difficulties in representing and estimating continuous output variables like activity time and duration. As traditional discrete choice methods can only work with categorical outputs, continuous outputs in utility-based models are often segmented into several groups to make them suitable for discrete choice. For example, activity start time or trip departure time were grouped into four or six separate periods \citep{daysim-bowman1998day, cemdap-bhat2004comprehensive}. This aggregate time representation reduces the accuracy of activity generation modules in utility-based models. To enhance discrete choice capacity, several studies have proposed different methods such as joint discrete-continuous models and hazard-based models to estimate continuous activity duration or time expenditure.

In contrast, rule-based models represent individual activity decisions as heuristics or rules \citep{lr-rasouli2014activity}. These models usually capture the activity patterns of individuals reasonably well. Their results may stem from the use of observed distributions in activity generation modules. These distributions are usually built from travel survey data. However, rule-based models might depend on complex manually-coded expert knowledge to generate activity patterns \citep{ml-drchal2019-data-driven}. This complexity may reduce the transferability of rule-based models, as these expert-designed components may differ among different cities or regions.

Machine learning is a practical approach to automatically derive rules from data. It could help to improve rule-based models by reducing the complexity in expert-designed components. Machine learning also has the potential to address the challenge of large datasets and new travel data sources \citep{lr-miller2017modeling}. Another advantage of machine learning is its systematic validation process, which separates data into training and validation sets \citep{lr-miller2017modeling}. Different machine learning methods are increasingly being applied in activity-travel behaviour research, where popular methods are neural networks and decision trees \citep{lr-koushik2020machine}. For activity generation tasks, for instance, simple neural networks were applied by \cite{nn-kato2002microsimulation} in a comprehensive activity-based modelling system. In addition, decision trees were used for generating activity attributes in activity-based models like ALBATROSS and DDAS \citep{albatross-arentze2000-albatross, ml-drchal2019-data-driven}. When summarising the performance of different machine learning methods for classification tasks in various travel behaviour studies, random forest usually delivers better results \citep{lr-koushik2020machine}.

The recent advancement of deep learning has led to breakthroughs in machine learning research \citep{goodfellow2016deep}. Deep learning is widely used to address various tasks in Natural language processing and Computer vision fields. In transportation research, many studies have focused on the application of deep learning for mode choice and traffic flow prediction \citep{lr-do2019survey, veres2019deep}. Due to the capacity in modelling nonlinearity, deep neural networks can predict traffic flow with high accuracy. Deep neural networks have recently been complemented with entity embedding to encode categorical features. Entity embedding technique not only helps to effectively encode categorical features, but also improves the accuracy of deep learning models \citep{guo2016entity, zheng2019deep}.

Till date, most research in activity-based modelling has focused on applying machine learning for mode choice and activity type detection \citep{lr-koushik2020machine}. However, there have not been many studies on incorporating machine learning into activity generation modules \citep{lr-koushik2020machine, ozonder2021longitudinal}. Moreover, the emergence of deep learning and new entity embedding techniques are yet to be explored for activity generation tasks. To address this gap, we develop in this paper a framework to apply deep learning and entity embedding for the improvement of the activity generation module. Our goal is to automatically learn the distributions from the observed survey data, and then use the learned distributions to generate activity and travel patterns. 

To this end, we introduce a novel activity pattern generator framework by leveraging the advantages of state-of-the-art machine learning methods including deep neural networks, random forest, and entity embedding techniques. Our contributions include:

\begin{itemize}
  \item Develop a novel activity pattern generator framework by complementing deep learning techniques with domain knowledge. The framework can effectively work with high-cardinality categorical features by incorporating entity embedding in deep neural networks.
  \item Replicate tour-based activity patterns by leveraging skeleton schedule knowledge. These tour-based travel patterns could play an important role in modelling travel mode choice constraints among activities.
  \item Propose a new approach that represents activity type as discrete, and both activity start time and end time as continuous variables. This approach can capture the activity start time and end time of the primary activity, as well as the start time pattern of stop-before and stop-after of the primary activity.
\end{itemize}
 
The rest of the paper is outlined as follows: \cref{literater-review} reviews existing approaches for activity generation tasks. \cref{act-gen} presents the proposed activity generator framework, from high-level architecture to detailed components. \cref{dataset} describes the data used to train and evaluate the model. Finally, \cref{discuss} discusses the results and important implications of our approach for developing the next generation of activity-based demand models.

\section{Literature Review}\label{literater-review}

In the following sections, we review related studies and approaches on activity generation tasks. We then discuss the application of machine learning to travel behaviour analysis. Furthermore, the combination of deep learning and entity embedding techniques in transportation research will also be presented.

Focusing on activity generation tasks, we categorise activity-based models into three groups: utility-based, rule-based and machine learning (ML-based) models. For each group, popular models will be discussed. \cref{table:temporal-resolution} summarises and compares the temporal resolutions and methods used for activity generation in these models. A more comprehensive review of different activity-based demand models can be referred to \cite{lr-rasouli2014activity}.

\begin{table}[h!]
      \centering
      \caption{Activity temporal resolution of activity generation modules in different activity-based models}
      \setlength\fboxsep{0pt}
      \vskip-\topsep%
      \smallskip%
      \renewcommand\arraystretch{1.2}
      \colorbox{darkgray}{%
      \footnotesize
      \begin{tabularx}{0.955\textwidth}{llllll}
      \toprule
Model name & Model category   & Start time & Generation method      & Duration & Generation method      \\
      \midrule
DaySim               & Utility-based    & Segments   & Discrete choice        &                   & None                   \\
CEMDAP               & Utility-based    & Segments   & Discrete choice        & Continuous         & Hazard-based model     \\
TASHA                & Rule-based       & Continuous  & Observed distributions & Continuous    & Observed distributions \\
ADAPTS               & Rule-based       & Continuous  & Observed distributions & Continuous         & Hazard-based model     \\
ALBATROSS            & ML-based 		& Segments   & Decision trees         & Segments          & Decision trees         \\
DDAS                 & ML-based 		&            & None                   & Continuous         & Decision trees   \\
       \bottomrule
       \end{tabularx}%
      }
      \label{table:temporal-resolution}
\end{table}

\subsection{Existing activity generation methods}
\subsubsection{Utility-based models}

One popular utility-based model is DaySim or Day Activity Schedule proposed by \cite{daysim-bowman2001activity}. They assumed that a prior basic activity pattern was formed before the decision of more detailed activity and travel plans. DaySim represented each individual schedule as a daily activity pattern with tours. The primary tour's structure included a primary activity, the sub-tour type (only for \textit{subsistence} primary activity), and intermediate stop-type before and after the primary activity location. \cite{daysim-bowman2001activity} used nested logit models to represent the choice of activity patterns. The highest tier of the nested hierarchy was an activity pattern, followed by a primary tour, and secondary tours at the lowest layer. The time of home-based tours was represented as the combination of the departure time from home and from the primary activity in the daily activity chain. These departure times were grouped into four time periods including AM peak, Midday, PM peak, and others \citep{daysim-bowman2001activity}. Two multinomial logit models were applied to estimate the choice of tour time for both primary and secondary tours. DaySim was then implemented as an activity generation component in an operational travel forecasting system named SacSim \citep{daysim-bowman2006sacramento}.

Another utility-based model is the Comprehensive Econometric Microsimulator for Daily Activity-Travel Patterns (CEMDAP) which formed an individual activity-travel structure into three levels including pattern, tour and stop \citep{cemdap-bhat2004comprehensive}. A pattern was a list of consecutive tours, and each tour included a chain of travel stops. Each travel stop was an out-of-home activity episode along with activity type, duration, and location, as well as the travel time to this stop location. CEMDAP differentiated travel patterns among non-workers, and workers (including students) \citep{cemdap-bhat2004comprehensive}. Non-worker patterns were represented with a list of home-based tours with no fixed stops. Worker patterns were divided into five segments: before-work, home-to-work, work-based, work-to-home, and after-work. For work activity, CEMDAP first generated work duration, and then start time. Specifically, a hazard-based model was used for generating continuous activity duration, while discrete choice methods were applied for selecting activity start time \citep{cemdap-bhat2004comprehensive}.

In utility-based models, discrete choice methods were normally used to model categorical outputs such as activity types and travel mode choices. To address the limitation of traditional discrete choice models when dealing with continuous activity time, different joint discrete-continuous choice models were proposed. For instance, \cite{bhat2005multiple} developed the multiple discrete-continuous extreme value (MDCEV) model for activity type and duration, while \cite{konduri2010probit} introduced a probit-based discrete-continuous model of activity type and duration choices. Furthermore, \cite{nurul2018comprehensive} proposed a comprehensive utility-based system by integrating discrete activity type and location choice with continuous-time expenditure for workers.

The advantages of utility-based models are their econometric-based travel behaviour theory and their capacity to represent activity schedules as tour-based travel patterns. The theory-driven modelling paradigm explains the behaviour of human choices, while tour-based patterns help to effectively forecast mode choices with constraints. However, the number of independent variables in these models is often small. In addition, many personal and household features are often ignored in utility functions. Despite some limitations in dealing with continuous outputs like activity start time, discrete choice methods have been extended for the development of different activity-based models due to their interpretability. For example, DaySim was integrated as the daily activity generation module for SimMobility-Midterm, as well as was extended in a weekly activity generation model named ActiTopp \citep{simmobi-lu2015simmobility, mobitopp-hilgert2017modeling}. 

\subsubsection{Rule-based models}

Two popular rule-based operational models are TASHA and ADAPTS. The Travel/Activity Scheduler for Household Agents (TASHA) was developed to replicate the activity scheduling and interacting process of household members \citep{tasha-miller2003prototype, tasha-miller2015implementation}. In TASHA, the sequence of activity generation was activity frequency, start time and duration. These activity attributes were generated from empirical probability distributions in a travel survey \citep{tasha-eberhard2003-act, tasha-miller2015implementation}. The formulation of observed distributions for different activity types was based on the cross-classification of individual, household and schedule characteristics such as gender, age, occupation, employment status, student status, presence of children, and work project status \citep{yasmin2015assessment}. The validation of TASHA's activity generation and scheduling modules showed that it can replicate observed activity with good accuracy \citep{tasha-roorda2008validation}. However, the formation of distributions was dependent on the demographic characteristics of studied areas  \citep{yasmin2015assessment}. Therefore, it remains challenging to transfer these observed rules into other cities.

\cite{adapts-auld2009-framework} developed the Agent-based Dynamic Activity Planning and Travel Scheduling (ADAPTS) model. ADAPTS aimed to dynamically model the activity planning and scheduling process. It treated the decision of selecting activity attributes as a separate event in each simulation time step. In any time step, if an agent planned an activity, the agent would firstly select an activity type. After that, an activity planning order module decided in what order other activity attributes were planned \citep{adapts-auld2009-framework}. For example, one planning order may be activity time first, followed by location, with whom, and finally by which travel mode. To model the dynamic planning order, ADAPTS required an additional travel survey called the Urban Travel Route and Activity Choice Survey, which may not be available in other cities \citep{adapts-auld2012-activity-planning}. Furthermore, similar to TASHA, activity start time and duration in ADAPTS were drawn from observed start time duration distributions derived from a travel survey \citep{adapts-auld2011agent}.

Both TASHA and ADAPTS have been implemented in operational agent-based systems \citep{tasha-miller2015implementation,adapts-auld2016polaris}. These systems may also incorporate discrete choice into other components such as mode choice modules. For activity generation modules, the use of observed distributions results in high accuracy of activity start time. However, the forming of these distributions might require expert domain knowledge. Machine learning methods can alleviate this limitation by automatically deriving rules from observed datasets, which will be explored in this paper.

\subsubsection{Machine learning (ML-based) models}
\vspace{3mm}
\textit{First generation neural networks}
\vspace{2mm}

The idea of applying neural networks for travel demand analysis could be traced back to the 1990s. The Activity-Mobility Simulator (AMOS) was probably the first attempt to use neural networks in a comprehensive activity-based travel demand system \citep{nn-kitamura1993amos}. AMOS was based on the behavioural principle of adaptation, which aimed to replicate a learning process where individuals try to find the best activity-travel alternative using a trial-and-error process. AMOS included a response option generator component which produced and ranked a list of choices when an individual encountered changes in their travel environment \citep{nn-kitamura1996sequenced}. Hence, the role of neural networks in AMOS was not for activity pattern generation, but for forecasting the change of activity pattern, given the change in the travel environment. 

In another study, \cite{nn-kato2002microsimulation} used neural networks to build an activity-based travel demand model for work-tour mode and related discretionary activities. They assumed the work travel pattern, particularly work travel mode, will affect the characteristics of discretionary tours before and after work. Firstly, a model was built to predict a travel mode to work. Given the work travel information, they then developed two models to estimate trip generation, destination choice, mode choice and activity duration for discretionary activities before and after work. This model showed its practical capability to forecast the impacts of travel demand measures on daily travel patterns \citep{nn-kato2002microsimulation}.

While the above studies showed the practical capability of neural networks for travel demand modelling, the architecture of these \textit{first-generation} neural networks was simple, using only one or two layers with a small number of neurons. The data sets used for model estimation were also small-size samples. Thus, these models did not fully exploit the advantages of neural networks.

\vspace{3mm}
\textit{Decision trees}
\vspace{2mm}

\cite{albatross-arentze2000-albatross} proposed A Learning-Based, Transportation-Oriented Simulation System (ALBATROSS), which was the first tree-based comprehensive activity-based model. ALBATROSS used different decision tree algorithms for generating activity attributes, including activity type, with whom, duration, start time and location. Activity start time was divided into six periods, while duration was grouped into three categories: short, average or long \citep{albatross-arentze2000using-dtree}. The alternatives of activity attributes relied on the choice situation which was based on different conditional variables. A limitation of ALBATROSS was its scheduling process which is based on a pre-assumption that is not validated on empirical data \citep{lr-rasouli2014activity}.

\cite{ml-drchal2019-data-driven} have recently developed a Data-Driven Activity Scheduler (DDAS) to generate sequential activity schedules. For activity generation, they assumed that previous activities affected the current activity. Hence, the current activity attributes such as activity type and end time were used to forecast the next activity attributes. \cite{ml-drchal2019-data-driven} employed a decision tree classifier to select the next activity type, and used a continuous probability distribution to generate the next activity duration. However, DDAS did not capture the start time of activities well \citep{ml-drchal2019-data-driven}. In addition, DDAS framework cannot produce tour-based activity schedules, which could reduce its accuracy in the mode choice module.

Nevertheless, the advantage of decision trees is their interpretability, which allows tree-based models such as ALBATROSS and DDAS to infer causal relationships among input factors and outputs. However, the accuracy of activity generation in these models seems weak, leading to more inaccurate outcomes in other modules such as location choice and mode choice modules.

Several studies have recently applied more advanced decision tree techniques such as random forest for trip generation \citep{ml-ghasri2017-data-mining} or activity generation \citep{ozonder2021longitudinal}. They showed the usefulness of random forest in producing higher accuracy for trip generation, as well as in identifying the most important explanatory variables for activity generation. Our proposed framework also exploits the advance of random forest for activity generation tasks.

\subsection{Deep neural networks for travel behaviour analysis}

Deep learning, or deep neural network, is a powerful tool for analysing not only images and text data but also tabular or relational data \citep{dl-arik2019tabnet}. With the improvement of high-performance computing systems and hardware like graphics processing units, machine learning researchers and practitioners are now able to train deep learning models with large datasets. Another advantage of deep learning is its ability in solving regression tasks, which can forecast continuous outputs. In travel behaviour analysis, there is an increasing application of machine learning, especially neural networks for predicting travel mode and travel destination \citep{lr-koushik2020machine}. However, there are not many studies on applying machine learning for activity generation tasks \citep{lr-koushik2020machine, ozonder2021longitudinal}. Especially, the integration of deep learning for developing a complete activity generation module is yet to be developed.

Deep learning has recently been complemented with entity embedding techniques. The use of entity embedding for encoding categorical features has improved the accuracy of deep neural networks and their capacity to work with categorical variables \citep{guo2016entity}. Entity embedding has a similar idea as word embedding like \textit{Word2Vec} encoding which represents each word as a continuous vector \citep{mikolov2013efficient}. Using entity embedding, which creates embedded vector representation for categorical variables, deep learning models can work well with categorical features with a large number of discrete values. Entity embedding helps to reduce the dimensions of categorical features, thus reducing the computation costs and speeding up neural networks compared with one-hot encoding\footnote{One-hot encoding represents each categorical value as a vector of a single one (1) and several zeros (0)} \citep{guo2016entity}. Moreover, the embedding vectors obtained from training deep neural networks can improve the performance of other machine learning methods like random forest. However, the combination of deep learning with entity embedding has only recently been applied in transportation research.

In the next section, we introduce a novel activity pattern generator that incorporates random forest, deep learning and entity embedding techniques to generate reliable mobility patterns.

\section{Proposed activity pattern generator}\label{act-gen}

In this section, we first introduce several concepts and the high-level architecture of the proposed framework, followed by more detailed components. Furthermore, the implementation of advanced machine learning techniques is described along with performance evaluation metrics.

\subsection{Tour-based activity pattern concept}\label{sec:act-ptt-concept}

We assume that each person's daily out-of-home activity schedule includes one primary activity and several secondary activities, as similar to \citep{daysim-bowman1998day}. The primary activity then forms a home-based primary tour, while secondary activities create several home-based secondary tours. The criteria to decide which activity is the primary activity will depend on activity type and activity duration. 

\vspace{8mm}
\textit{Activity type and activity group}
\vspace{2mm}

Activity types are categorised into three groups: \textit{subsistence}, \textit{maintenance}, and \textit{discretionary} (\cref{table:act-cat}). As activity type and activity group have a similar role, they can be used interchangeably. \textit{Subsistence} activities include work for workers and study for students. \textit{Maintenance} activities are those for the household to maintain their daily life. Shopping and personal business, for instance, are maintenance activities. \textit{Discretionary} activities are those with lower priority and constraints. For example, \textit{discretionary} activities include recreational, social, and other activities. However, different from \citep{daysim-bowman1998day}, we explicitly consider \textit{pickup-dropoff} activities as a separate activity group. While \textit{pickup-dropoff} activities normally take a short duration, they hold important information regarding individuals travel patterns. 

\begin{table}[h!]
      \centering
      \caption{Activity group along with its activity types}
      \setlength\fboxsep{0pt}
      \vskip-\topsep%
      \smallskip%
      \renewcommand\arraystretch{1.2}
      \colorbox{darkgray}{%
      \begin{tabularx}{0.55\textwidth}{lcl}
      \toprule
      Activity group		& Code	& Activity type     \\
      \midrule
      Subsistence			& W		& Work; Study             \\
      Maintenance			& M 	& Shopping; Personal business \\
      Discretionary			& D 	& Recreation; Social; Other   \\
      Pickup–dropoff		& P 	& Pick up or Drop off    \\
       \bottomrule
       \end{tabularx}%
      }
      \label{table:act-cat}
\end{table}

\vspace{3mm}
\textit{Primary and secondary activity}
\vspace{2mm}

We define a primary activity and secondary activities for three different personal groups: workers, students, and nonworkers (\cref{table:act-prim-second}). For workers and students, the primary activity could be \textit{subsistence}, \textit{maintenance}, or \textit{discretionary}; while secondary activities could be \textit{maintenance}, \textit{discretionary}, or \textit{pickup-dropoff}. For nonworkers, since they do not have a \textit{subsistence} activity, their primary activity could be \textit{maintenance} or \textit{discretionary}. Nonworkers also have similar secondary activity types like workers and students. 

Several conditions with thresholds are used to decide which is a primary activity. The thresholds depend on activity groups and the dataset's characteristics. The highest priority is for \textit{subsistence} activity, followed by \textit{maintenance} (with a minimum duration above a threshold), and then \textit{discretionary} (again with a minimum duration above a threshold). When no \textit{subsistence} primary activity is found, we consider a primary activity as the longest duration one among \textit{maintenance} and \textit{discretionary} activities in that person's activity schedule.

\begin{table}[h!]
      \centering
      \caption{Primary activity and secondary activity in different personal groups}
      \setlength\fboxsep{0pt}
      \vskip-\topsep%
      \smallskip%
      \renewcommand\arraystretch{1.2}
      \colorbox{darkgray}{%
      \begin{tabularx}{0.58\textwidth}{lcc}
      \toprule
      Personal group		& Primary activity	& Secondary activity \\
      \midrule
      Workers				& W or M or D	& M or D or P \\
      Students				& W or M or D 	& M or D or P \\
      Nonworkers			& M or D		& M or D or P \\   
       \bottomrule
       \end{tabularx}%
      }
      \label{table:act-prim-second}
\end{table}

\vspace{3mm}
\textit{Formulation of primary tour and secondary tour}
\vspace{2mm}

\begin{figure}[!h]
  \centering
  \includegraphics[scale=0.80]{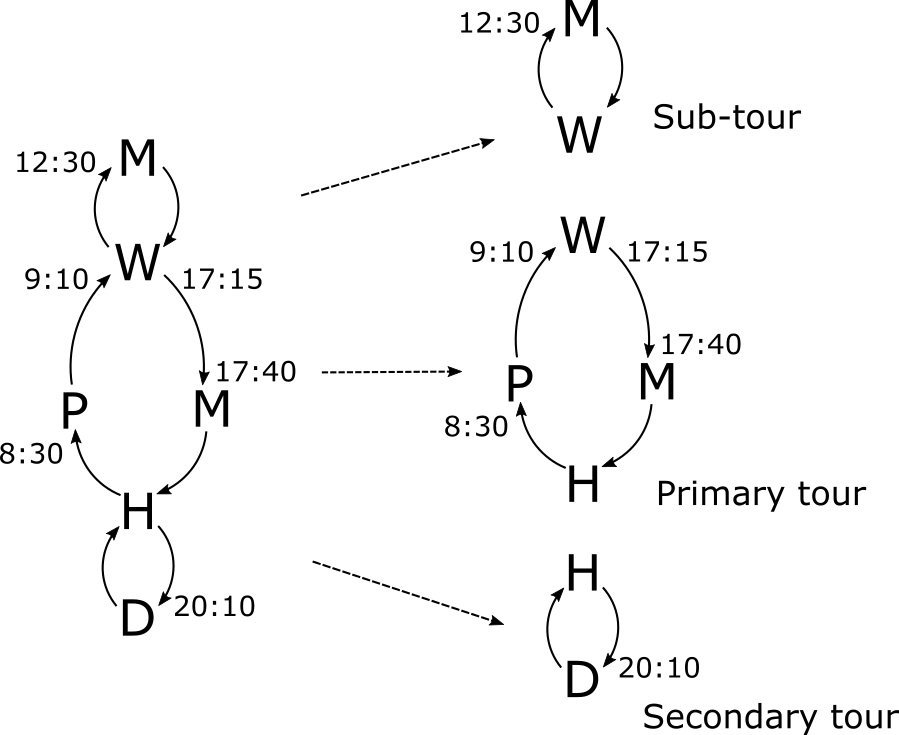}
  \caption{Translating an activity schedule to tour-based activity patterns}
  \label{fig:tour-based-pattern}
\end{figure}

Each activity schedule is represented as a home-based primary tour along with several home-based secondary tours. The primary tour may have stop-before or stop-after or both. The primary tour with \textit{subsistence} primary activity might contain sub-tours. These concepts are illustrated in \cref{fig:tour-based-pattern} where an activity schedule is transformed into one home-based primary tour (with a work-based sub-tour) and one home-based secondary tour.

For example, one exemplary daily activity schedule of a worker is described as follows: In the morning, the worker dropped off (P) their child at 8:30 am on the way to a workplace. The worker then started to work (W) at 9:10 am, and at noon, went out for a personal business (M) at 12:30 pm for around one hour, and after that returned to work. The worker finished working at 5:15 pm and went shopping (M) on the way home. The person arrived at a shopping centre at 5:40 pm and spent around 30 minutes shopping. The worker then came home, rested, and had dinner with the family. At night, the person went out to a nearby cafe to socialise (D) with their friends for one hour, and then came back home. From this information, which can be derived from a travel diary survey, we can construct an activity schedule for that worker (as in the left of \cref{fig:tour-based-pattern}).

The activity schedule is translated into one primary tour and one secondary tour (as in the right of \cref{fig:tour-based-pattern}). The primary tour has work (W) as a primary activity. There is one stop-before (P), and one stop-after (M) the primary activity. In addition, the person also has a work-based sub-tour (M) at noon during work activity. The transformation from activity schedule into a primary tour and secondary tours helps to form the architecture of the proposed activity generator framework.

\subsection{High-level architecture}\label{sec:High-level architecture}

\begin{figure}[!h]
  \centering
  \includegraphics[scale=0.8]{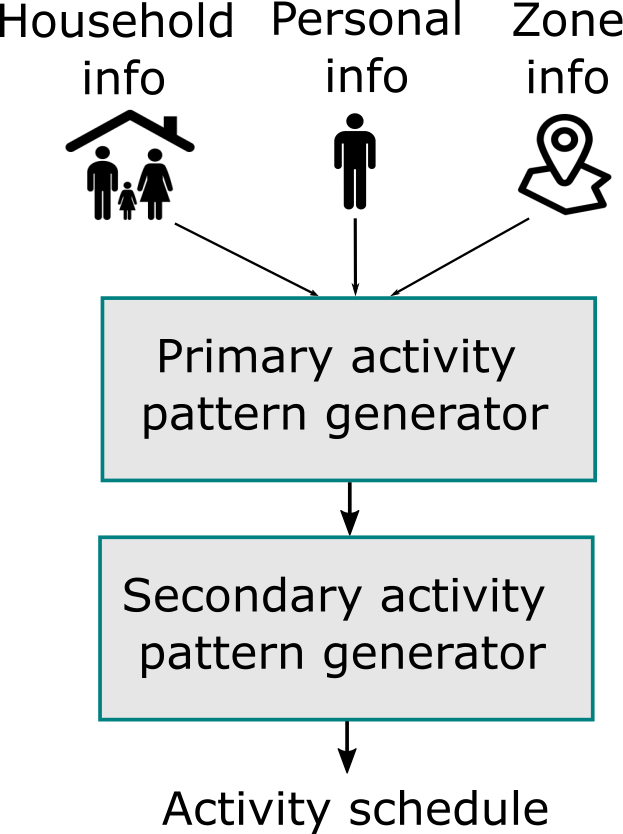}
  \caption{High-level architecture}
  \label{fig:architecture}
\end{figure}

The activity pattern generator framework aims to generate individual tour-based activity schedule (\cref{fig:architecture}). The inputs include household demographics, personal characteristics, and zone information. The framework first generates primary activity attributes which are considered as the skeleton schedule of each person. Based on the skeleton information, the framework then produces a full primary activity pattern of stop-before, stop-after, and sub-tour attributes. Finally, the framework creates secondary tour-based activities based on the primary activity pattern information.

In the probability terms, each activity schedule is represented as a set of secondary home-based tours tied together by a primary activity pattern (\cref{eq:sched}). The primary activity pattern is a home-based primary activity tour which includes primary activity attributes along with stop-before and stop-after it (\cref{eq:primPat}).

\begin{equation}\label{eq:sched}
	prob(schedule) = prob(primaryPattern) \cdot prob(secondaryPattern|primaryPattern)
\end{equation}

\begin{equation}\label{eq:primPat}
	prob(primaryPattern) = prob(primaryActivity) \cdot prob(stops|primaryActivity)
\end{equation}

The assumption here is that the primary activity affects other activities and stops in a daily schedule of individuals. The primary activity forms a skeleton pattern that constraints stop-before, stop-after, and secondary activities. The benefit of this formulation is that it could generate tour-based travel patterns, which could yield more reliable outcomes in the subsequent modules such as location choice and mode choice of activity-based models. In contrast, models like DDAS can only generate the current activity based on the previous activity information, without constraints on tours \citep{ml-drchal2019-data-driven}. Specifically, the mode choice module of DDAS is a trip-based model. Compared to trip-based models, tour-based models deliver a major advance as they explicitly account for the logical interconnections between individual trips \citep{miller2019agent}.

To this end, our proposed activity pattern generator framework includes two modules: a primary activity pattern generator and a secondary activity generator (\cref{fig:architecture}). The framework first uses the primary pattern generator to generate individuals primary tour-based activity patterns. Then, based on the primary activity information, the framework will generate secondary tour-based activities using the secondary pattern generator. In the current setup, the secondary activity generator component is simpler than the primary activity generator. Similar to \cite{tasha-miller2003prototype}, the sequence in predicting secondary activity attributes is activity type, and then start time and duration. Detailed components and implemented methods are presented in the next sections.

\subsection{Primary activity pattern generator}\label{sec:prim-pattern-gen}

\begin{figure}[!h]
  \centering
  \includegraphics[scale=0.8]{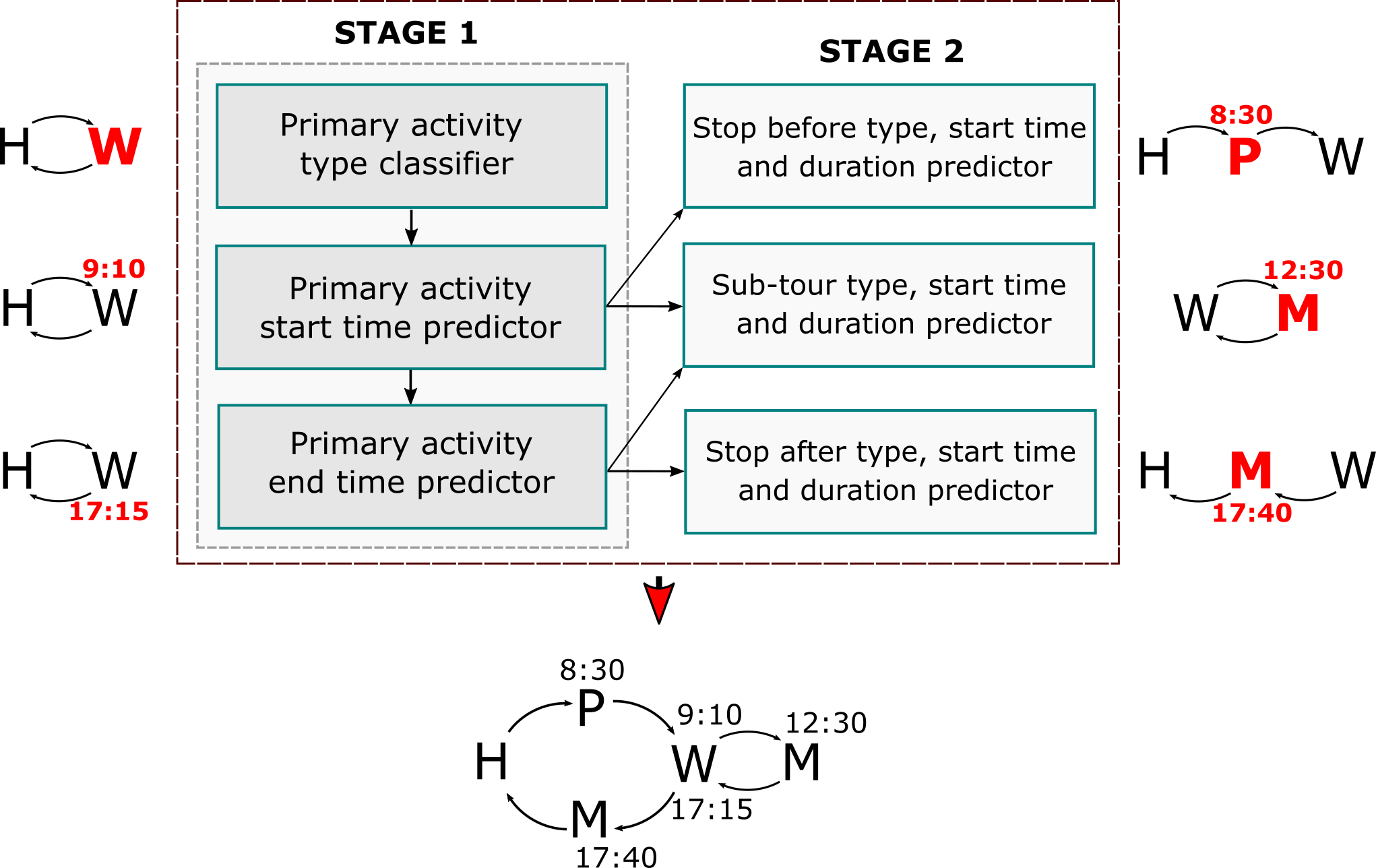}
  \caption{The detailed architecture of primary activity pattern generator. The bottom shows an example of final primary activity patterns for the daily scheduler mentioned in \cref{fig:tour-based-pattern}.
		\\Stage 1: Generating primary activity attributes, with the (red) output of each step on the left.
		\\Stage 2: Generating stops and sub-tour information, with the (red) output of each step on the right.
  }
  \label{fig:prim-architecture}
\end{figure}

A primary activity pattern generator module is used to generate a home-based primary tour. This process includes two main stages: first generating primary activity attributes, and then generating stop-before, stop-after, and sub-tour for this primary activity (\cref{fig:prim-architecture}). 

\vspace{2mm}
\textit{Primary activity attributes}

In the first stage, we classify the primary activity type as one of \textit{subsistence}, \textit{maintenance} or \textit{discretionary} categories. Given the primary activity type, the model then predicts its continuous start time. After that, the model will estimate the end time of the primary activity based on its type, start time, and other inputs. The primary duration will be calculated from the gap between the start time and end time. The output of stage one is the primary activity type, its start time and end time.

It is worth noting that in contrast to other existing models, we predict the end time of the primary activity instead of its duration. This is because there may be a sub-tour during the primary activity. In this case, using the end time of the primary activity represents more accurate activity time patterns. When a sub-tour occurs, the duration of primary activity can be derived from the differences in primary activity's end time and start time, less the duration of the sub-tour.

\vspace{2mm}
\textit{Stop-before and stop-after of a primary activity}

The second stage produces the stop-before and stop-after of the primary activity. Stop information is generated based on its primary attributes, which can be considered as a skeleton schedule. Given the primary activity type, and start time, the model predicts the type, start time and duration of the stop-before of a primary activity. The activity type of stop-before can be \textit{maintenance}, \textit{discretionary}, \textit{pickup-dropoff}, as well as \textit{zero-stop} type in case there is no stop. Similarly, the model also forecasts the stop-after given the primary activity type and end time.

This second stage could also generate sub-tour information if the primary activity type is work or education. The sub-tour activity type can be \textit{maintenance}, \textit{pickup-dropoff}, \textit{discretionary} or \textit{None}. In this situation, the sub-tour attributes are generated using the primary activity type, and both the start time and end time of the primary activity.

\subsection{Machine learning techniques}\label{sec:ml-technqs}

We develop different deep neural networks and random forest models for each component in the primary activity generator and secondary activity generator. Classification models are implemented for classifying stop or activity types, while regression models are built for activity time prediction.

\vspace{3mm}
\textit{Random forest}
\vspace{2mm}

Random forest (RF) is a combination of several decision trees $ ({h(x, \Theta_{k}), k = 1,...}) $, where the $ \Theta_{k} $ represents independent identically distributed random vectors, and each tree decides a weight for the most common features of the input $ x $ \citep{rf-breiman2001random}. Random forest can be applied to sort the feature importance of input variables in both classification tasks (like activity type classification) and regression tasks (like activity start time prediction) \citep{rf-breiman2001random}.

\begin{figure}[!h]
  \centering
\includegraphics[scale=0.45]{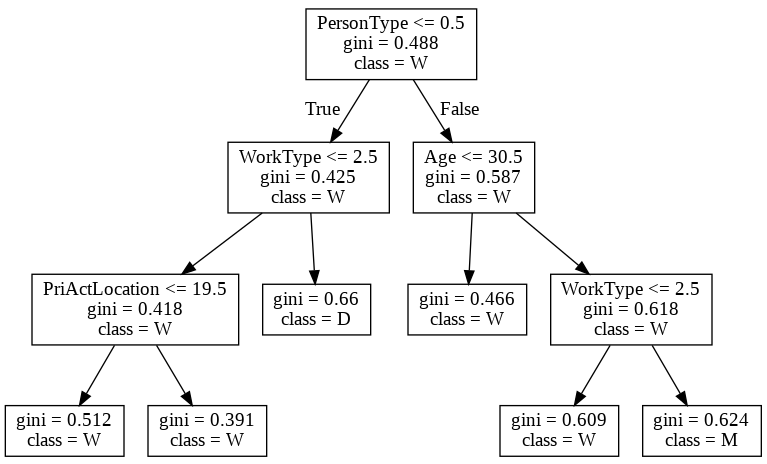}
  \caption{A decision tree for workers' primary activity type classifier}
  \label{fig:decision_tree}
\end{figure}

Each decision tree has nodes and leaves. The top nodes represent independent features, while the bottom contains tree leaves, which represent final targets. For example, one simple decision tree for the workers' primary activity type classification is shown in \cref{fig:decision_tree}. This tree has five nodes with variable names (MainRole, WorkType, Age, and PriActLocation) along with conditions to divide it into branches. At the bottom, there are six leaf nodes, which show the predicted label of primary activity types (W: \textit{Work}, M: \textit{Maintenance}, or D: \textit{Discretionary}).

A grid search algorithm\footnote{https://scikit-learn.org/stable/modules/generated/sklearn.model\_selection.GridSearchCV.html} with five-fold cross-validation is used to find the best random forest configuration of hyperparameters (\cref{table:hype-params-rf}). For simplicity, we investigate the performance of different random forest models by varying the number of decision trees, and the minimum of sample leaves in each decision tree. The Gini index is used to measure the node impurity for classifiers, while the residual sum of squares metric is applied for regression models \citep{liaw2012randomforest}. The max samples depend on the size of the training sets. For example, there is a maximum of 10000 samples in case of workers' data set, while there is a maximum of 5000 samples in each decision tree for students and nonworkers' data set since the number of observations is smaller. The investigated configurations of different random forest models are detailed in \cref{table:hype-params-rf}. In addition, the importance of variables is also calculated to examine the explanatory variables in random forest models.

\begin{table}[h!]
      \centering
      \caption{Random forests' hyperparameters}
      \setlength\fboxsep{0pt}
      \vskip-\topsep%
      \smallskip%
      \renewcommand\arraystretch{1.2}
      \colorbox{darkgray}{%
      \begin{tabularx}{0.38\textwidth}{lcc}
      \toprule
      Hyperparameter		& & Values  \\
      \midrule
      max\_samples			& & 5000, 10000 		\\
      min\_samples\_leaf	& & 3, 5, 6, 10, 20 	\\
      n\_estimators			& & 40, 60, 100	\\
       \bottomrule
       \end{tabularx}%
      }
      \label{table:hype-params-rf}
\end{table}

\vspace{3mm}
\textit{Deep neural networks with entity embedding}
\vspace{2mm}

We use Deep feedforward network architecture, also called Multilayer perceptrons (MLPs), to build deep neural networks. Feedforward neural networks are a conceptual stepping stone for the development of other architectures like recurrent networks (RNNs) and convolutional networks (CNNs) \citep{goodfellow2016deep}. We choose MLP architecture due to its simplicity and suitability for tabular data which is a travel survey as in our case.

MLPs are the composition of many different hidden layers \citep{goodfellow2016deep}. Each hidden layer is represented as a function $h$, which is the combination of an activate function $g$ on top of an affine transformation (\cref{eq:hid}). The affine transformation is used to transform the input $(x)$ from the previous layer by using parameters $(W)$ and bias $(b)$. The activate function is a nonlinear transformation. Popular activate functions include $Sigmoid$, $Tanh$, Rectified linear unit ($ReLU$), and $Leaky ReLU$. For $n$ hidden network layers, the output function $f$ is represented in \cref{eq:f_func} with the input $h^{(n)}$. The $h^{(n)}$ function is calculated from its previous hidden layers $h^{(n-1)}$ as presented in \cref{eq:h_func}.

\begin{equation}\label{eq:hid}
	h = g(W^{T}x + b)
\end{equation}

\begin{equation}\label{eq:f_func}
	f = g^{(n)}(W^{(n)T}h^{(n)} + b^{(n)})
\end{equation}

\begin{equation}\label{eq:h_func}
	h^{(n)} = g^{(n-1)}(W^{(n-1)T}h^{(n-1)} + b^{(n-1)})
\end{equation}

When implementing different deep neural networks for different components in the activity pattern generator, we also use a simple grid search approach to seek the best architecture of feedforward neural networks. We vary the number of hidden layers from one to six layers, and find that two or three hidden layers produce the best results. The configuration of other hyperparameters is presented in \cref{table:hype-params-dl}.

\begin{table}[h!]
      \centering
      \caption{Deep neural networks' hyperparameters}
      \setlength\fboxsep{0pt}
      \vskip-\topsep%
      \smallskip%
      \renewcommand\arraystretch{1.2}
      \colorbox{darkgray}{%
      \begin{tabularx}{0.60\textwidth}{lcc}
      \toprule
      Hyperparameter			& & Values  \\
      \midrule
      number of hidden layers	& & 1, 2, 3, 4, 5, 6 \\
      learning rate				& & $1e^{-5}$, $1e^{-4}$, $1e^{-3}$, $1e^{-2}$, $1e^{-1}$\\
      activation				& & $Sigmoid$, $ReLU$ \\
      loss function				& & Cross entropy, RMSE	\\
      optimiser					& & SGD, Adam \\
      \bottomrule
       \end{tabularx}%
      }
      \label{table:hype-params-dl}
\end{table}

\begin{figure}[!h]
  \centering
\includegraphics[scale=1]{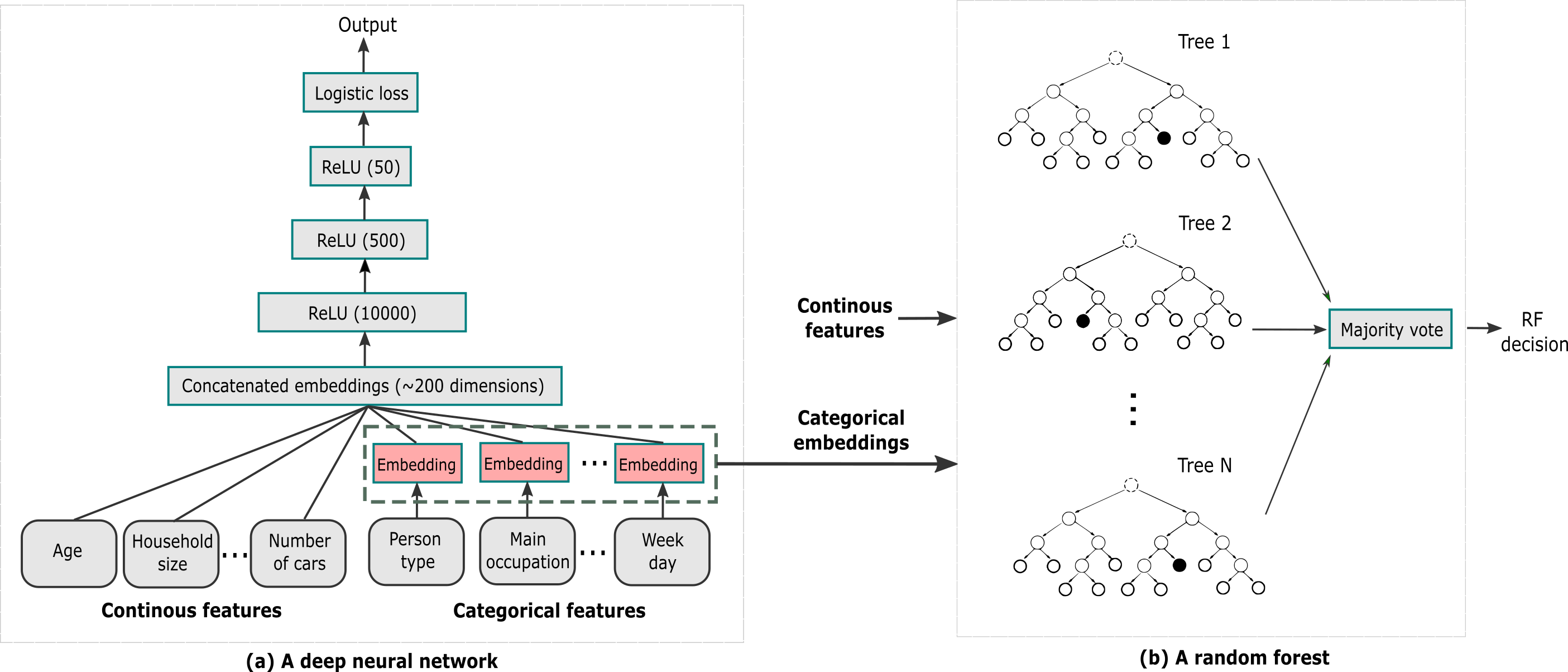}
  \caption{Transferring categorical embedding vectors from a deep neural network to a random forest.
		\\(a) A deep neural network with 3 hidden layers and 1 entity embedding layer for categorical features.
		\\(b) A random forest model which uses the categorical encoded vectors from the deep neural network. 
  }
  \label{fig:embedding-randomforest}
\end{figure}

The above MLPs architecture is complemented with the entity embedding of high-cardinality categorical
features to form our activity pattern generator framework. We apply entity embedding to encode categorical variables of the household, individual and zone information.  These encoded features include variables such as person type, main occupation, weekday and home postcode. The encoded vectors are then concatenated with continuous variables like person age, household size, and the number of cars. For example, a deep neural network architecture for primary activity start time prediction includes three hidden layers with $ReLU$ activation, as well as an entity embedding layer for categorical variables (as shown in (a) of \cref{fig:embedding-randomforest}). 

One-hot encoding is a popular technique to encode categorical variables in Machine Learning. Compared to one-hot encoding, entity embedding requires less memory (see \cref{table:emb-weekday} for an example). With entity embedding, for instance, a categorical variable with $n$ unique values is encoded by a vector size of $ \frac{n + 1}{2} $ in our setup, while one-hot encoding requires a vector size of $n$. For instance, \cref{table:emb-weekday} shows the difference between one-hot encoded vector and entity embedded vector for a weekday, which is derived from the deep neural network for classifying the workers' primary activity type.

To investigate the efficacy of the entity embedding technique, we compare the performance of deep neural networks with one-hot encoding and with entity embedding. The learned entity embedded vectors are also integrated into RF models to check its transferability by copying the learned embedded vectors from a deep neural network to a random forest model, as shown in \cref{fig:embedding-randomforest}. We use Scikit-learn\footnote{https://scikit-learn.org/} library for building the random forest models, as well as Pytorch\footnote{https://pytorch.org/} and Fastai\footnote{https://www.fast.ai/} libraries for implementing the deep neural networks with different embedding techniques.

\begin{table}[h!]
      \centering
      \caption{An example of one-hot encoding and entity embedding representation for weekday.}
      \setlength\fboxsep{0pt}
      \vskip-\topsep%
      \smallskip%
      \renewcommand\arraystretch{1.2}
      \colorbox{darkgray}{%
      \begin{tabularx}{0.60\textwidth}{lcc}
      \toprule
      Weekday			& One-hot encoding	& Entity embedding  \\
      \midrule
      Monday			& [1, 0, 0, 0, 0, 0, 0]		& [0.56, -0.87, 0.57, -0.76] \\
      Tuesday			& [0, 1, 0, 0, 0, 0, 0] 	& [-0.73, -1.33, 0.29, -1.68] \\
      Wednesday			& [0, 0, 1, 0, 0, 0, 0] 	& [0.09, -1.42, -0.48, 0.81] \\
      Thursday			& [0, 0, 0, 1, 0, 0, 0] 	& [0.14, 0.22, 0.21, -1.82] \\
      Friday			& [0, 0, 0, 0, 1, 0, 0] 	& [-0.74, -0.80, 1.22, 0.03] \\
      Saturday			& [0, 0, 0, 0, 0, 1, 0] 	& [-1.04, 2.45, 0.14, -0.56] \\
      Sunday			& [0, 0, 0, 0, 0, 0, 1] 	& [0.27, 0.89, -0.56, 1.58] \\
       \bottomrule
       \end{tabularx}%
      }
      \label{table:emb-weekday}
\end{table}

\vspace{3mm}
\textit{Model evaluation metrics}
\vspace{2mm}

The $accuracy$ rate (\cref{eq:acc}) and $F1\_score$ (\cref{eq:f1}) is used to measure the fitness of classification models like activity type classification, while Root Mean Square Error $(RMSE)$ is used to measure the error of regression models (\cref{eq:rmse}) for activity time prediction. These metrics will be compared for the results of different deep learning (DL) and random forest (RF) models, with and without entity embedding. 

\begin{equation}\label{eq:acc}
	Accuracy \footnote{ TP: True Positives, TN: True Negatives, FP: False Positives, FN: False Negatives } = \frac{TP+TN}{TP+FP+TN+FN}
\end{equation}

\begin{equation}\label{eq:f1}
	F1\_score = 2 \cdot \frac{\mathrm{Precision} \cdot \mathrm{Recall}}{\mathrm{Precision} + \mathrm{Recall}} = \frac{TP}{TP + 0.5 \cdot (FP + FN)}
\end{equation}

\begin{equation}\label{eq:rmse}
	RMSE = \sqrt{(\frac{1}{n})\sum_{i=1}^{n}(y_{i} - \hat{y}_{i})^{2}}
\end{equation}

\section{Data}\label{dataset}

The Victorian Integrated Survey of Travel and Activity (VISTA)\footnote{https://transport.vic.gov.au/about/data-and-research/vista} is used for the framework's implementation. This travel diary survey includes around 174,000 trips of 64,500 persons in 25,000 households from Victoria, Australia from 2012 to 2018. The survey's variables include personal variables, household features and zone information. Detailed description of these features are presented in \cref{table:variable_desc}.

A data cleaning process is performed and after removing inconsistent rows, there are around 158,000 trips of 41,700 persons in 19,600 households, which are used for model estimation and validation as described below. These trips are then converted into 41,700 personal activity schedules including 23,900 workers, 9,900 students (from primary schools to universities), and 7,900 nonworkers.

\begin{table}[h!]
      \centering
      \caption{The description of the dataset's variables}
      \small
      \setlength\fboxsep{0pt}
      \vskip-\topsep%
      \smallskip%
      \renewcommand\arraystretch{1.1}
      \colorbox{darkgray}{%
      \begin{tabularx}{\linewidth}{l l c X} 
      \toprule
Variable           & Data type        & Range/\#Values & Description                                                                        \\
      \midrule 
Personal attributes &                  &                 &                                                                                    \\
		\hline
PersonType          & Categorical      & 3               & Worker; Student; Nonworker                                                           \\
Age                 & Numerical (int)  & {[}0-116{]}     & Age of the person                                                                  \\
Gender              & Categorical      & 2               & Male; Female                                                                                \\
CarLicence          & Categorical      & 2               & Yes; No                                                                                \\
MainOccupation      & Categorical      & 10              & Main occupation                                                                    \\
ANZSCO2             & Categorical      & 53              & Detailed occupation                                                                \\
MainIndustry        & Categorical      & 21              & Main industry                                                                      \\
ANZSIC2             & Categorical      & 106             & Detailed industry                                                                  \\
PersonIncomeLevel   & Categorical      & 11              & Income level                                                                       \\
MainRole            & Categorical      & 7               & Full-time worker; Part-time worker; Student; Pupil; Child; Retired; Nonworker         \\
WorkType            & Categorical      & 5               & Fixed hours; Flexible hours; Roster shifts; Work from home; Not in workforce    \\
EmpType             & Categorical      & 5               & Employment type                                                                    \\
		\hline
Household features  &                  &                 &                                                                                    \\
		\hline
OwnDwell            & Categorical      & 5               & Fully owned; Being purchased; Being rented; Occupied rent-free; and Something else \\
TravelYear          & Numerical (int)  & {[}2012-2018{]} & Year travel, from 2012 to 2018                                                     \\
TravelMonth         & Numerical (int)  & {[}1-12{]}      & Travel month \\
TravelDay           & Categorical      & 7               & Day in week                                                                        \\
NumPersons          & Numerical (int)  & {[}1-11{]}      & Number of persons in a household                                                    \\
NumKids             & Numerical (int)  & {[}0-7{]}       & Number of children in a household                                                       \\
NumFulltimeWorkers  & Numerical (int)  & {[}0-6{]}       & Number of full-time workers in a household                                           \\
NumParttimeWorkers  & Numerical (int)  & {[}0-5{]}       & Number of part-time workers in a household                                           \\
NumCasualeWorkers   & Numerical (int)  & {[}0-4{]}       & Number of casual workers in a household                                            \\
NumCars             & Numerical (int)  & {[}0-7{]}       & Number of cars                                                                     \\
NumBikes            & Numerical (int)  & {[}0-14{]}      & Number of bikes                                                                    \\
HhIncome            & Numerical (cont) & {[}0-12500{]}   & Household income                                                                   \\
YearsLived          & Numerical (int)  & {[}0-88{]}      & Number of years lived at the house                                                 \\
		\hline
Zone information    &                  &                 &                                                                                    \\
		\hline
HomeRegion          & Categorical      & 2               & Home region                                                                        \\
HomeLGA             & Categorical      & 32              & Home local government area                                                   \\
HomePostcode        & Categorical      & 245             & Home postcode                                                                     
\\	  
      \bottomrule
      \end{tabularx}%
      }
      \label{table:variable_desc}
\end{table}

Individual trips are translated into activity attributes which then form each person's activity schedule as a list of consecutive activities (as described in \cref{sec:act-ptt-concept}). For each person, the primary activity is derived from the activity schedule based on activity priority. Similar to \citep{daysim-bowman1998day}, we define thresholds with several conditions to select primary activities. From the primary activity and activity schedule, a primary activity pattern is formed by combining primary activity attributes, stop-before, and stop-after of the primary activity (as explained in \cref{fig:tour-based-pattern}). Similarly, secondary activities are those in the activity schedule, but not in the primary activity home-based tour.

Data analysis shows that more than 96\% of people have \textit{zero-stop} or one stop-before or one stop-after each primary activity. Hence, to simplify the model implementation, we only consider the maximum of one stop-before and one stop-after each primary activity. In case a person has two or more stop-before or stop-after, we select the longest duration stop, and remove other stops. The number of secondary tours is also insignificant; thus it is omitted in the current model's implementation.

Finally, the data is split into training and validation sets. The data from 2012 to 2017 is used for training models, while the 2018 data is kept as validation sets.

The results of the activity generator framework, which incorporates machine learning techniques proposed in \cref{sec:ml-technqs} for both training and validation sets, will be presented in the next section.

\section{Results and discussion}\label{discuss}
\vspace{3mm}
\textit{Activity type classification}
\vspace{2mm}

In most cases, random forest (RF) performs better than deep learning (DL) in the primary activity type classification task (\cref{table:priacttype}). RF also mostly produces better accuracy rates and $F1\_score$ than DL in validation sets. This outcome is consistent with the results in previous studies \citep{lr-koushik2020machine}. Both RF and DL deliver reasonable accuracy for workers and students; however, the accuracy outcomes of classifying activity type in validation sets are weak. This is understandable, since the decision of selecting a primary activity type may depend on other factors outside the information in the dataset. For example, the decision not to travel to work should be based on the workers' weekly work schedule. Similarly, the decision to go to work on a specific day of a part-time worker should depend on their roster.

For nonworkers, however, the accuracy is weak in both training and validation sets for RF and DL (\cref{table:priacttype}). This may be due to the diversity and uncertainty of activity type selection of nonworkers. As they do not have to participate in \textit{subsistence} activities like work or school, there are fewer constraints in their decision making for primary activity type selection. Thus, there is a need for more details such as a weekly travel survey to get a comprehensive picture of activity schedules.

The integration of entity embedding helps to improve machine learning models. Combining entity embedding into deep neural networks mostly increases the accuracy and $F1\_score$ in both training and validation sets (on DL\&Emb columns of \cref{table:priacttype}). The embedded vectors, which are trained from DL, can also enhance the performance of RF models. Entity embedding not only helps RF to produce better outcomes in training sets, but it may also help RF deliver better results in validation sets (on RF\&Emb columns of \cref{table:priacttype}). This is due to the entity embedding's capacity to capture the intrinsic properties of categorical variables by mapping similar values close to each other in the embedding space \citep{guo2016entity}. Entity embedding helps deep neural networks and random forest models to generalise better. Hence, entity embedding provides a useful technique when dealing with datasets with many high cardinality features such as postcodes and detailed occupations.

\begin{table}[h!]
      \centering
      \caption{The result of the primary activity type classification (Emb: Entity embedding)}
      \setlength\fboxsep{0pt}
      \vskip-\topsep%
      \smallskip%
      \renewcommand\arraystretch{1.15}
      \colorbox{darkgray}{%
      \begin{tabularx}{1.05\textwidth}{lcccccccccc}
      \toprule
      \multirow{2}{*}{Group}& \multirow{2}{*}{Metric} 
      &\multicolumn{4}{c}{Training set} & &  \multicolumn{4}{c}{Validation set}\\
      \cline{3-6} \cline{8-11} 
      &    & DL	& DL{\footnotesize \&}Emb & RF    & RF{\footnotesize \&}Emb & 
      & DL & DL{\footnotesize \&}Emb & RF    & RF{\footnotesize \&}Emb \\
      \midrule 
Workers    & Acc & 0.724 & 0.754 & 0.804 & \textbf{0.820} & & 0.687 & 0.704 & \textbf{0.712} & 0.705 \\
Students   & Acc & 0.749 & 0.837 & 0.845 & \textbf{0.848} & & 0.780 & 0.798 & 0.835 & \textbf{0.841} \\
Nonworkers & Acc & \textbf{0.703} & 0.627 & 0.625 & 0.668 & & 0.584 & 0.594 & 0.576 & \textbf{0.594} \\

\hline
Workers    & F1 & 0.784 & 0.789 & 0.822 & \textbf{0.836} & & \textbf{0.759} & 0.753 & 0.752 & 0.747 \\
Students   & F1 & 0.779 & 0.853 & 0.872 & \textbf{0.873} & & 0.819 & 0.813 & 0.873 & \textbf{0.875} \\
Nonworkers & F1 & 0.711 & 0.668 & 0.693 & \textbf{0.714} & & 0.616 & 0.624 & \textbf{0.665} & 0.662 \\  

      \bottomrule
      \end{tabularx}%
      }
      \label{table:priacttype}
\end{table}

\begin{figure}[!h]
\centering
  \includegraphics[scale=0.55]{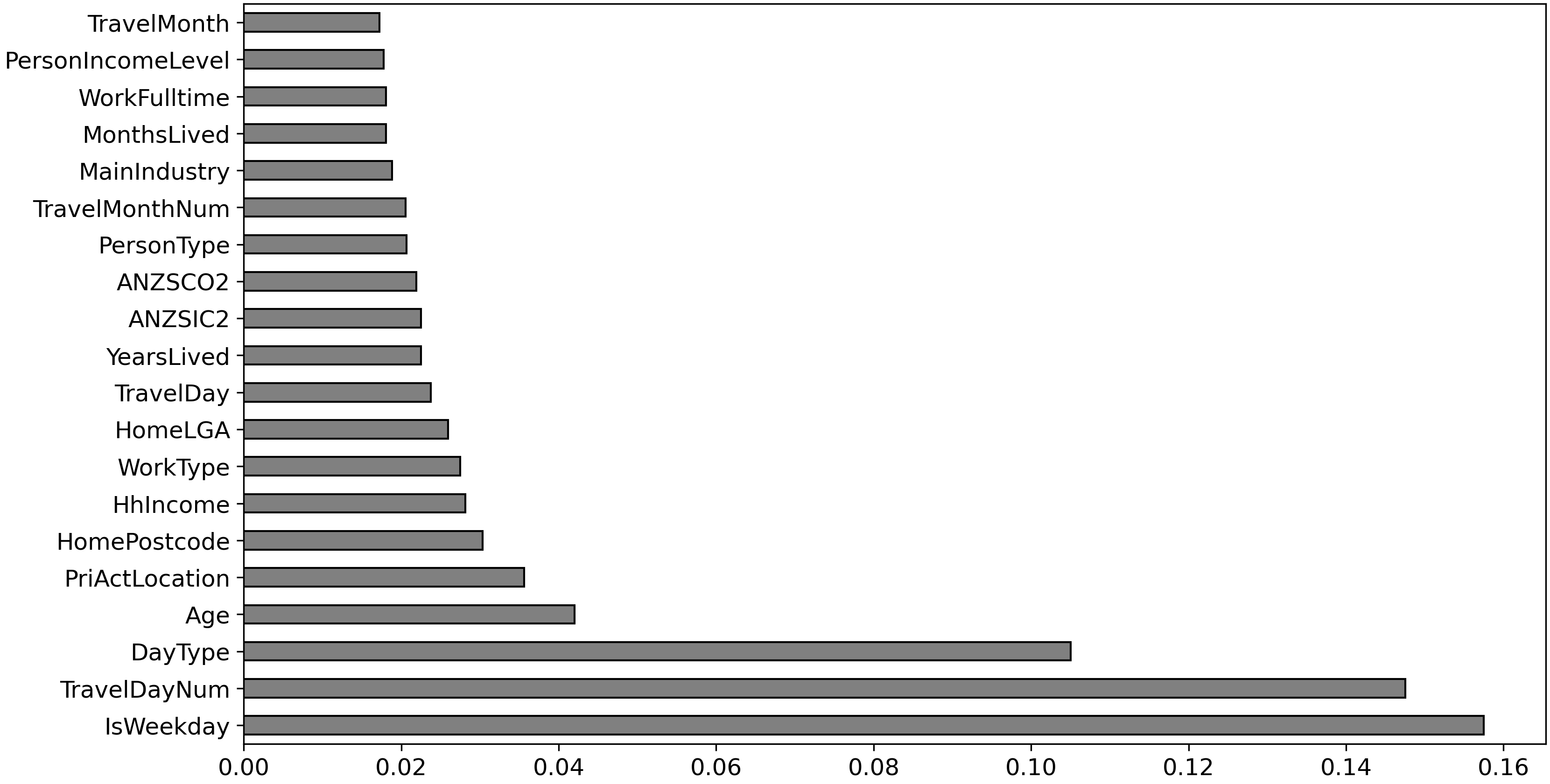}  
  \caption{Feature importance from a RF primary activity classifier}
  \label{fig:feature_importance}
\end{figure}

Compared to DL, an advantage of RF is that it can show how it makes forecasting decision via feature importance. This helps to improve the interpretability of our framework. \cref{fig:feature_importance} shows the top 20 features that most affect the primary activity classifier. The most influenced factors include activity location, home postcode, age, income, occupation, industry, and travel month.

\vspace{3mm}
\textit{Activity time prediction and primary activity patterns}
\vspace{2mm}

\begin{figure}[!h]
\centering
  \includegraphics[scale=0.5]{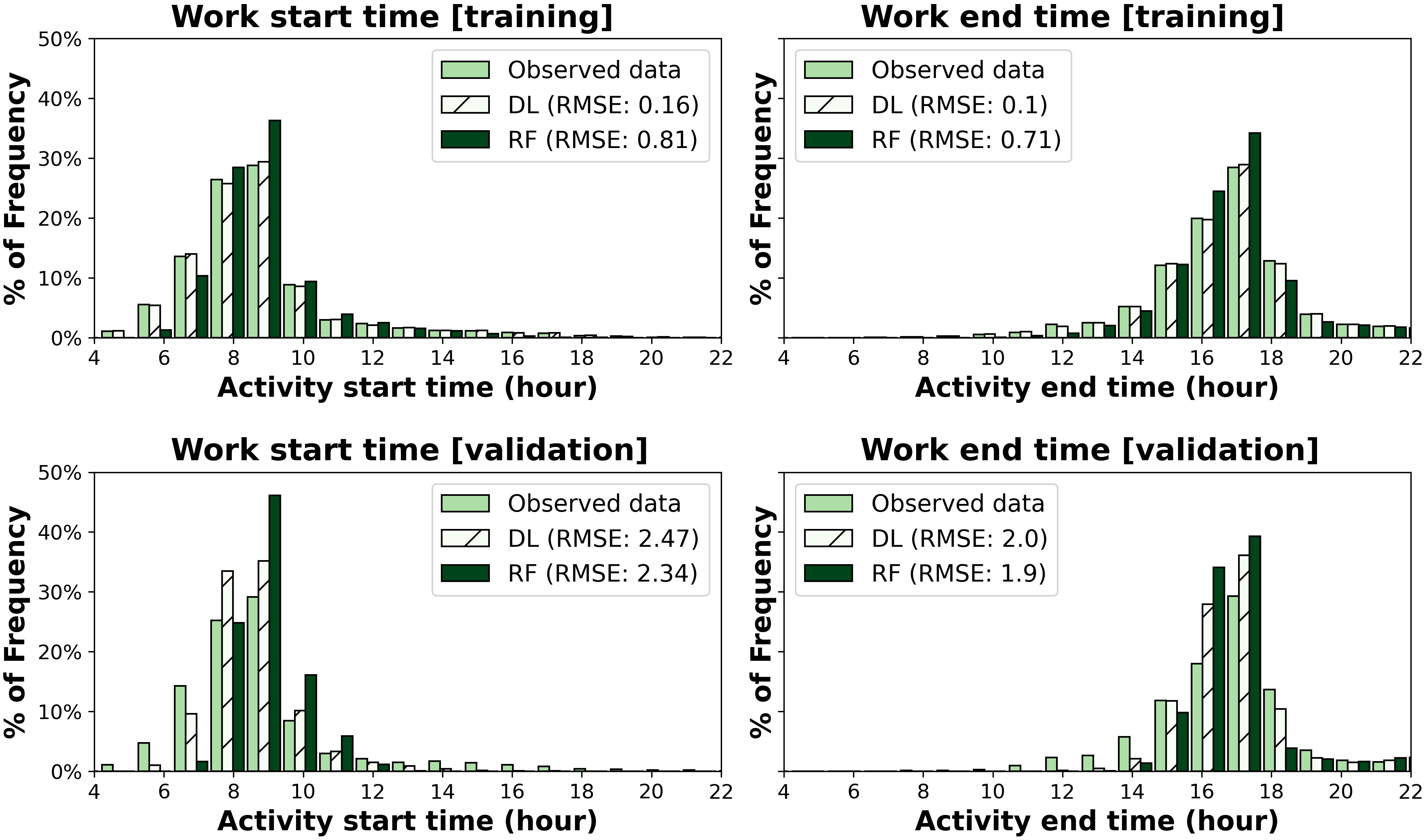}
  \caption{The start time and end time of work activities}
  \label{fig:worker-work-time}
\end{figure}

\begin{figure}[!h]
\centering
  \includegraphics[scale=0.5]{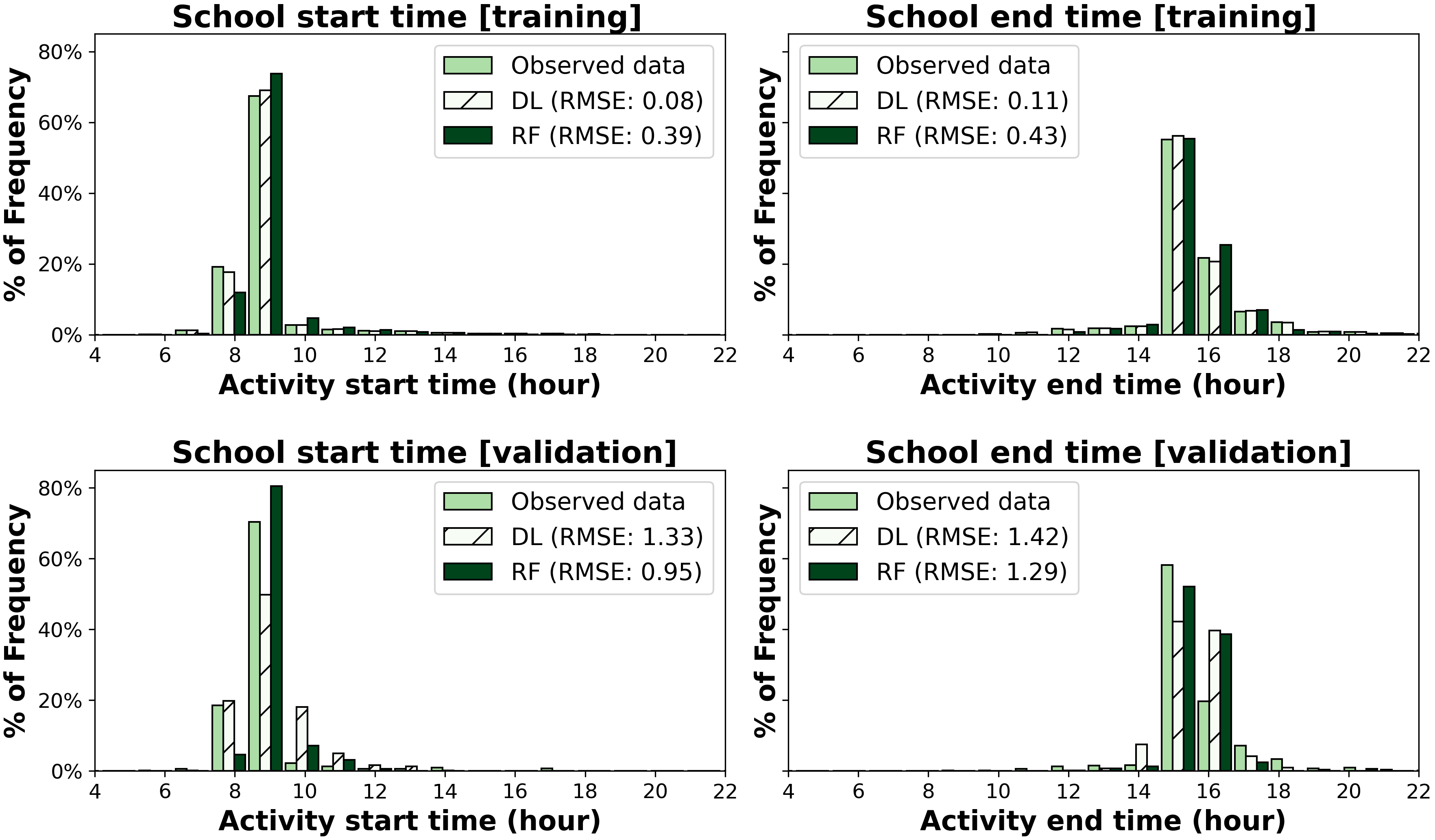}
  \caption{The start time and end time of school activities}
  \label{fig:student-school-time}
\end{figure}

\begin{figure}[!h]
  \centering
\includegraphics[scale=0.5]{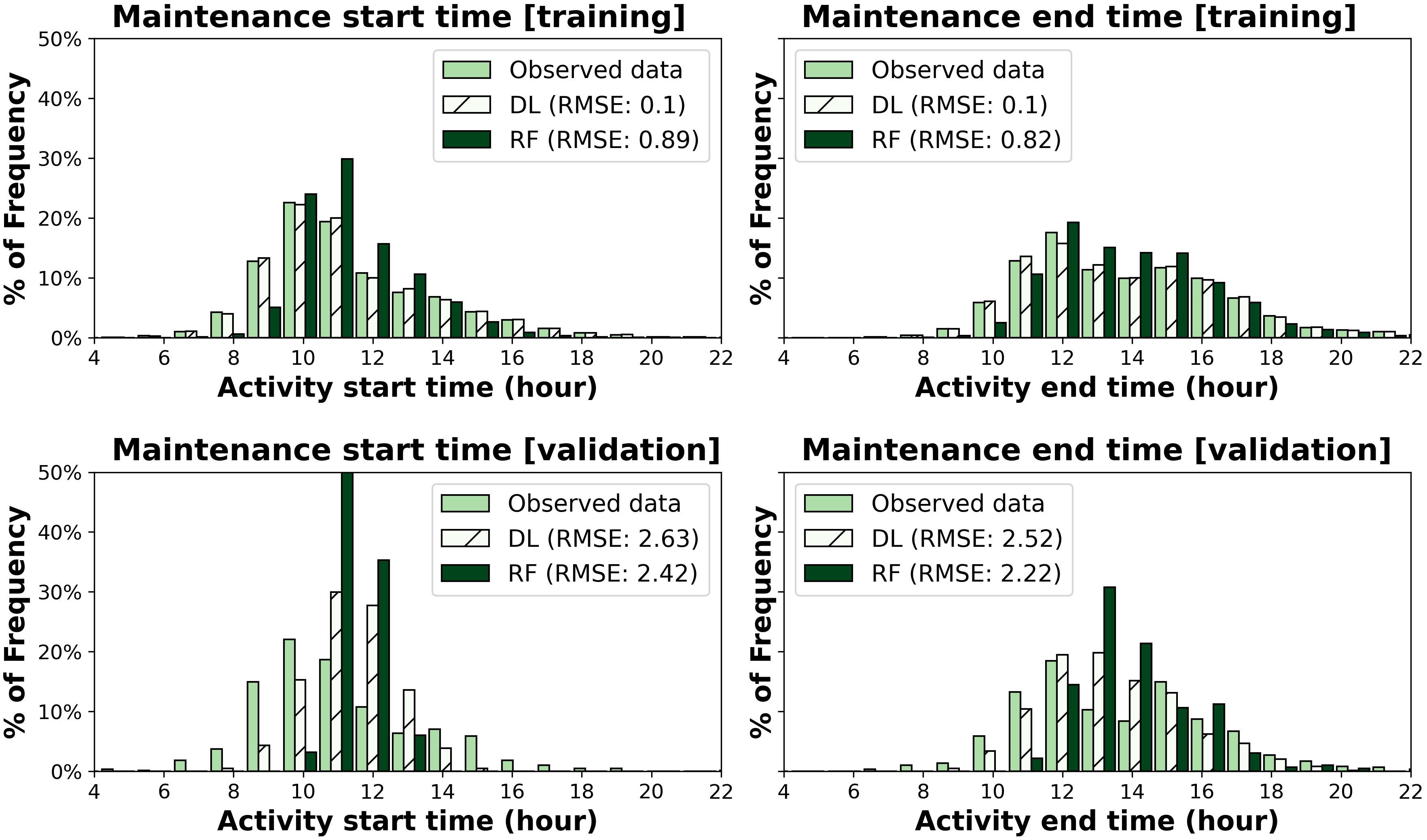}
  \caption{The \textit{maintenance} start time and end time of nonworkers}
  \label{fig:nonworker-mainten-time}
\end{figure}

\begin{figure}[!h]
  \centering
\includegraphics[scale=0.5]{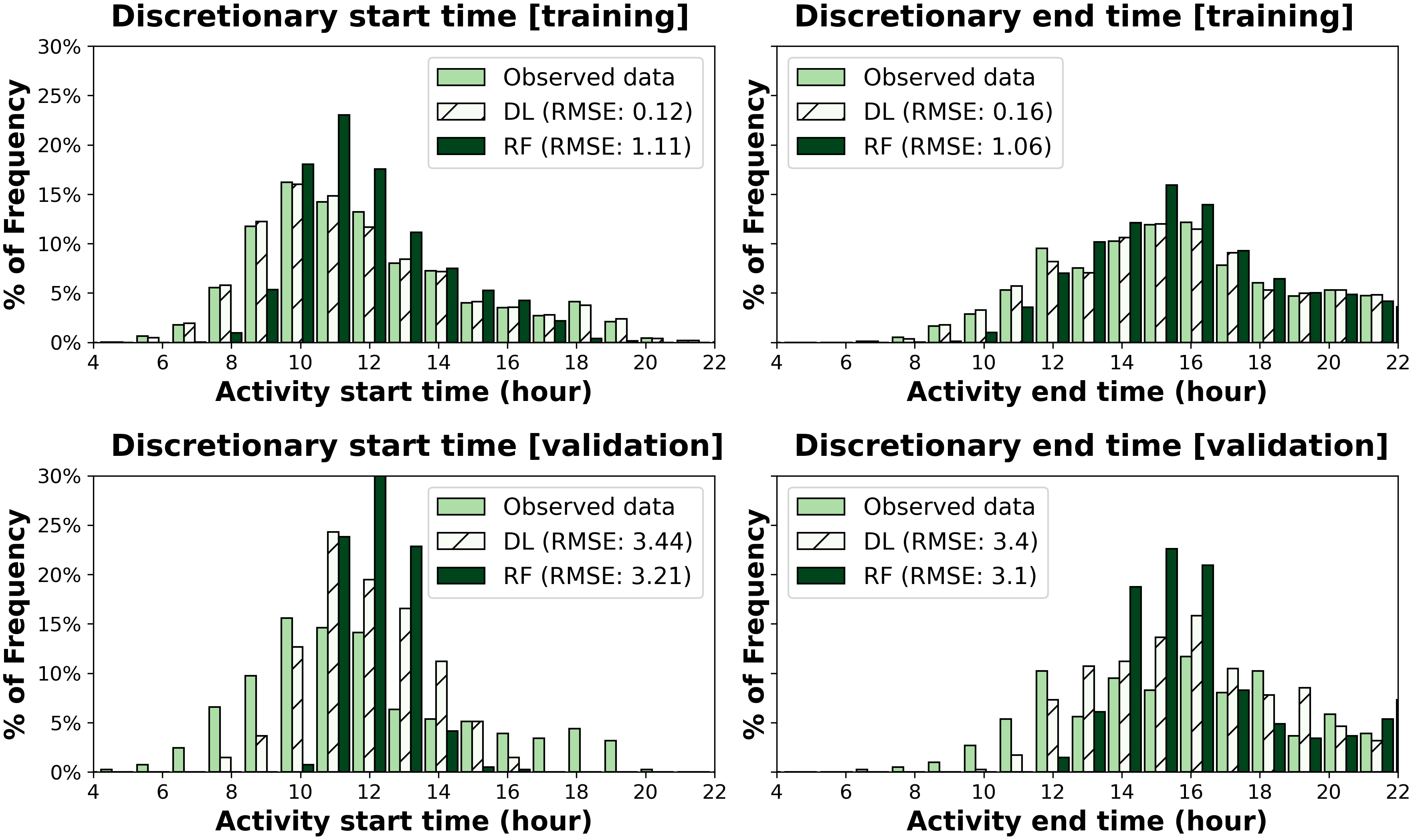}
  \caption{The \textit{discretionary} start time and end time of nonworkers}
  \label{fig:nonworker-discre-time}
\end{figure}

For primary activity start time and end time prediction, DL outperforms RF in training sets but is less generalised than RF in validation sets. DL can effectively replicate observed activity time of work activity for workers (\cref{fig:worker-work-time}) and school activity for students (\cref{fig:student-school-time}). However, the prediction errors in validation sets for all three groups (workers, students, nonworkers) are significantly higher than training sets. Especially, the patterns of activity start time for \textit{maintenance} and \textit{discretionary} activities are not captured well in validation sets (\cref{fig:nonworker-mainten-time}, \cref{fig:nonworker-discre-time}). RF seems less overfitting than DL in both training and validation sets.

\begin{figure}[!h]
  \centering
  \includegraphics[scale=0.5]{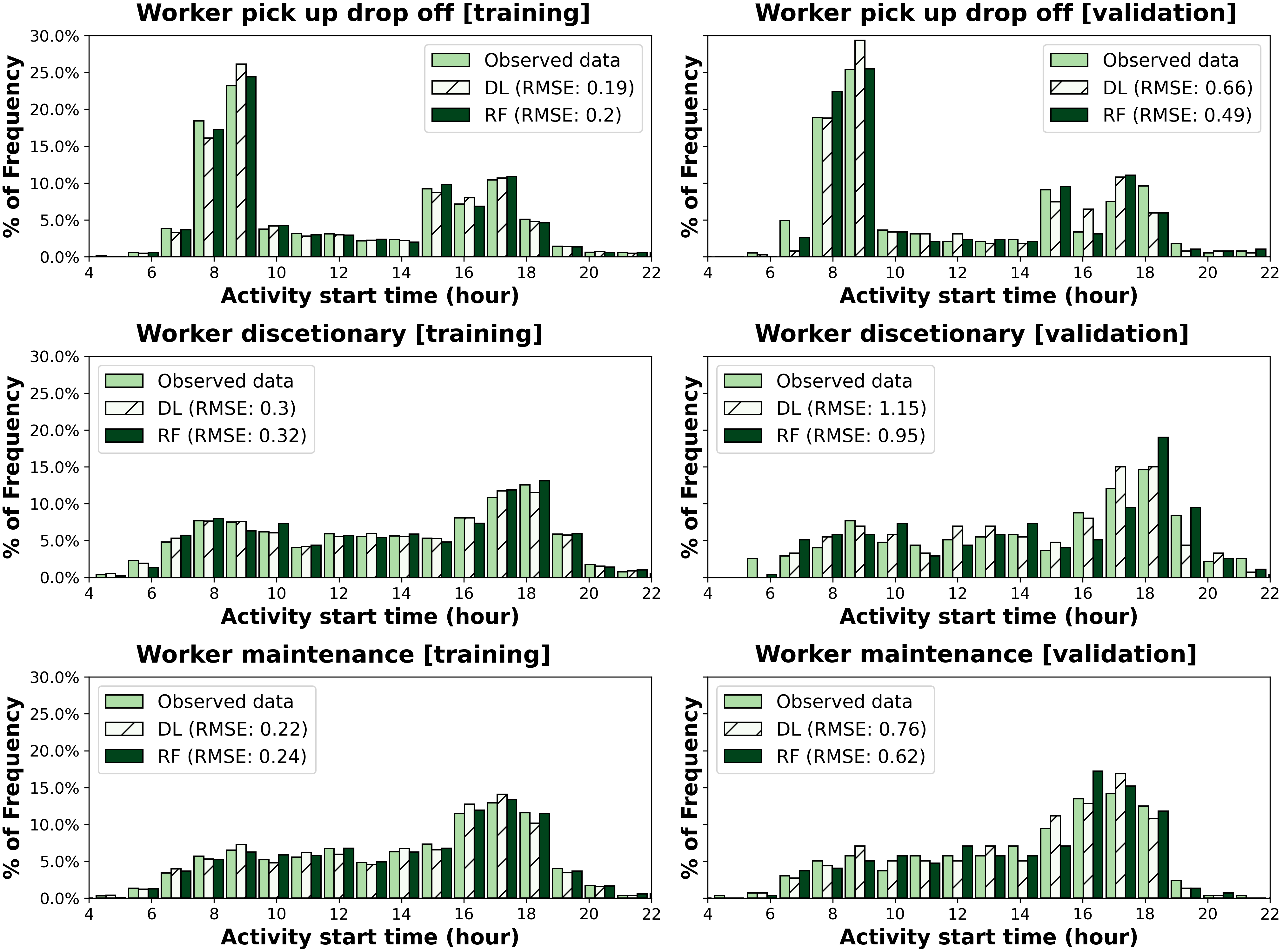}
  \caption{The start time of stop-before and stop-after of work activities}
  \label{fig:worker-stops}
\end{figure}

In addition, the framework accurately predicts the start time pattern of stop-before and stop-after the primary activity. Given the primary skeleton information, both DL and RF in the framework can capture the travel pattern before and after the primary activity well. For example, our model can effectively replicate the pattern of observed data for the \textit{pickup–dropoff}, stop-before, and stop-after of work activity (\cref{fig:worker-stops}). Compared to observations, both DL and RF produce similar patterns of activity start time throughout the day, especially the trend of peak hours in the morning and afternoon. Our approach produces more accurate patterns for nonwork/school activities in comparison with discrete choice methods which tend to over-predict the frequency of nonwork/school activities \citep{dianat2020modeling}.

\vspace{15mm}
\textit{Important implications}
\vspace{2mm}

The results show the potential use of deep learning and entity embedding in activity generation tasks. Our proposed activity generator framework accounts for a large number of explanatory features. This can help to address the challenge of big data and new travel behaviour data sources such as GPS, smartcard transactions, and smartphone applications. Machine learning methods, especially deep learning, can automatically estimate large numbers of parameters. Furthermore, combining with entity embedding, our framework can also effectively represent categorical features with a large number of discrete values. We also show that entity embedding has the potential to encode context-dependent variables to produce better outcomes for both deep neural network and random forest.

\vspace{5mm}
We demonstrate the need for incorporating travel behaviour domain knowledge with deep learning. While experimenting with different architecture to leverage deep learning, the use of the tour-based skeleton schedule in our framework helps to capture accurate primary tour patterns. Besides, the tour-based activity patterns could benefit mode choice modules by leveraging tour-based mode choice models. Deep learning is advantageous in forecasting continuous activity attributes such as start time, end time, and duration, while random forest could be suitable for activity type classification. Moreover, the discrete choice could be used to model variables with limited discrete values like travel modes. Hence, using domain knowledge to design appropriate model architecture that can exploit the advantages of both machine learning methods and discrete choice models.

In future work, the accuracy of the primary activity type classification can be further improved. For example, the use of a weekly travel survey could help to derive which days individuals go to work. Given the available weekly travel data, our framework can incorporate this information for its improvement.

\section{Conclusion}\label{conclusion}

We show that deep learning has the potential for activity generation tasks in travel demand systems, which perform well for primary activity time prediction in the proposed framework. This is underpinned by the deep learning's capacity in dealing with high-cardinality categorical inputs as well as continuous outputs prediction in large datasets. 
Combining with skeleton schedule knowledge, our approach can generate reliable activity patterns. More importantly, we also show that deep learning with entity embedding can accurately capture and reproduce the activity pattern for stop-before and stop-after of work primary activity. 
The framework could be expanded for activity location prediction by combining with additional data sets. It also provides a viable approach to exploit advanced machine learning techniques for generating more reliable activity and travel patterns, resulting in more reliable personal activity schedules and better accuracy performance for activity-based and agent-based transport systems.

\vspace{5mm}
\textbf{Acknowledgement}
\vspace{2mm}

The authors thank the Victorian Department of Transport, Australia for VISTA data access. We are also grateful for the feedback from Prof. Eric J. Miller and his research group at the Department of Civil \& Mineral Engineering, University of Toronto, Canada.

\normalsize
\bibliography{references}

\end{document}